\documentclass[]{spie}  %>>> use for US letter paper
\usepackage[table]{xcolor} %for use in color links 

\usepackage{multirow}
\usepackage[]{graphicx}

\usepackage{epsfig}
\usepackage{epstopdf}
\usepackage[]{todonotes}
\usepackage{booktabs}									% "schönere" Tabellen, größerer Abstand bei Linien

\usepackage{sidecap}

\usepackage[ruled]{algorithm2e}

\usepackage{amsmath}
\usepackage{url}

\usepackage{lscape} % flip table

\usepackage{tikz}
\usetikzlibrary{shapes}
\usetikzlibrary{arrows}
\usetikzlibrary{positioning}
\usetikzlibrary{mindmap}
\usetikzlibrary{fit}					% fitting shapes to coordinates
\usetikzlibrary{backgrounds}	% drawing the background after the foreground

\usepackage{pgfplots,pgfplotstable}

\usetikzlibrary{pgfplots.groupplots}

\title{An Evaluation of Trajectory Prediction Approaches and Notes on the {TrajNet} Benchmark.}

\author{Stefan Becker \footnote{\text{   }Equal contribution.} , Ronny Hug \footnotemark[\value{footnote}] , Wolfgang H\"{u}bner and Michael Arens
\skiplinehalf
Fraunhofer Institute for Optronics, System Technologies, and Image Exploitation IOSB\\
Gutleuthausstr. 1, 76275 Ettlingen, Germany
}
 
\begin{document} 
\maketitle

%%%%%%%%%%%%%%%%%%%%%%%%%%%%%%%%%%%%%%%%%%%%%%%%%%%%%%%%%%%%% 
\begin{abstract}
In recent years, there is a shift from modeling the tracking problem based on Bayesian formulation towards using deep neural networks. Towards this end, in this paper the effectiveness of various deep neural networks for predicting future pedestrian paths are evaluated. The analyzed deep networks solely rely, like in the traditional approaches, on observed tracklets without human-human interaction information. The evaluation is done on the publicly available \emph{TrajNet} benchmark dataset \cite{TrajNet_2018}, which builds up a repository of considerable and popular datasets for trajectory-based activity forecasting. We show that a Recurrent-Encoder with a Dense layer stacked on top, referred to as RED-predictor, is able to achieve sophisticated results compared to elaborated models in such scenarios. Further, we investigate failure cases and give explanations for observed phenomena and give some recommendations for overcoming demonstrated shortcomings. 
\end{abstract}

%%%%%%%%%%%%%%%%%%%%%%%%%%%%%%%%%%%%%%%%%%%%%%%%%%%%%%%%%%%%%
\section{INTRODUCTION}
\label{sec:intro} 
The prediction of possible future paths is a central building block for an automated risk assessment. The applications cover a wide range from mobile robot navigation, including autonomous driving, smart video surveillance to object tracking. Dividing the many variants of forecasting approaches can be roughly done by asking how the problem is addressed or what kind of information is provided. Firstly, addressing this problem reaches from traditional approaches like the Kalman filter \cite{kalman1960}, linear \cite{McCullagh_book_1989} or Gaussian regression models \cite{Williams1998}, auto-regressive models \cite{Hirotugu_1969}, time-series analysis \cite{Priestley_book_1981} to optimal control theory \cite{Kitani_ECCV_2012}, deep learning combined with game theory \cite{Ma_CVPR_2017}, or the application of deep convolutional networks \cite{Huang_TIP_2016} and recurrent neural networks (RNNs) as a sequence generation problem \cite{alahi2016social,alahi2017learning,Hug_RFMI_2017}. Secondly, the grouping can be done by using the provided information. On the one hand, the approaches can solely rely on observations of consecutive positions extracted by visual tracking or on the other hand, by using richer context information. This can be for example human-human interactions or human-space interactions or general additional visual extracted information like pedestrian head orientation \cite{Kooij_ECCV_2014} or head poses \cite{Hasan_CVPR_2018}. For some representative approaches which model human-human interactions, one should mention the works of Helbing and Moln\'ar \cite{Helbing_PRE_1995} and Coscia et al. \cite{Coscia_IVC_2018} or approaches in combination with RNNs like \cite{alahi2016social,alahi2017learning}. The spatial context of motion can in principle be learned by training a model on observed positions of a particular scene, but it is not guaranteed that the model successfully captures spatial points of interest and does not only implicitly keep spatial information by performing path integration in order to predict new positions. Nevertheless, here we distinguish such approaches from approaches where scene context is provided as further cue for example by semantic labeling \cite{Ballan_ECCV_2016} or scene encoding \cite{Xue_WACV_2018}. 
The challenges of \emph{Trajectory Forecasting Benchmarking} (\emph{TrajNet} $2018$) \cite{TrajNet_2018} are designed to cover some inherent properties of human motion in crowded scenes. The \emph{World H-H TrajNet} challenge in particular looks at predicting motions in world plane coordinates of human-human interactions. The aim of this paper is to find an effective baseline predictor only based on the partial history and find the maximum potential achievable prediction accuracy for this challenge. Achieving this objective involves an evaluation of different deep neural networks for trajectory prediction and analysis of the datasets properties. Further, we propose small changes and pre-processing steps to modify a standard RNN prediction model to result in a simple but effective RNN architecture that obtains comparable performance to more elaborated models, which additionally captures the interpersonal aspect of human-human interaction.\\
The paper is structured as follows. Firstly, the properties of the \emph{TrajNet} benchmark dataset are analyzed in section \ref{sec:data_ana}. Then, some basic deep neural networks are shortly described and evaluated. Further, the modifications in order to increase the prediction performance are presented (section \ref{sec:eval}). 
 The achieved results and an additional failure analysis are presented in section \ref{sec:discussion}. Finally, a conclusion is given in section \ref{sec:conclusion}. 

\section{{TrajNet} Benchmark Dataset Analysis}
\label{sec:data_ana}
%\todo[color=green!40]{Hier geht's los}
The trajectory forecasting challenges \emph{TrajNet} \cite{TrajNet_2018} provide the community with a defined and repeatable way of comparing path prediction approaches as well as a common platform for discussions in the field. In this section some properties of the current repository for the \emph{World H-H TRAJ} challenge of popular datasets for trajectory-based activity forecasting are analyzed and thereby design choices for the proposed predictor are deduced.\\

\begin{table}[h!]
\caption{Training (green) and test (cyan) dataset of the world plane human-human dataset challenge (adapted from the \emph{TrajNet} website \cite{TrajNet_2018}).}
\label{tab:dataset_detail}
  %\centering	
	%\rowcolors{2}{blue!5}{gray!10}
\def\arraystretch{1.1}	
\begin{tabular}{ |c| c c c c |}
%\begin{tabular}{p{2.5cm}l  c p{1cm} c p{2cm} c p{1cm} c p{1cm} c p{2cm}}
		\hline
					%\multicolumn{7}{c}{ Settings: $\sigma^{2}_{w}=2$, $\sigma^{2}_{r}=5$, $t_{update}=3$  }\\
					%\hline
			\rowcolor{white!10} 
		\multicolumn{1}{|c|}{Name} & \multicolumn{1}{c}{Resolution} & \multicolumn{1}{c}{\#Pedestrian }  & \multicolumn{1}{c}{Framerate} & \multicolumn{1}{c|}{Reference} \\
					\hline
				\rowcolor{green!10} 
				\emph{BIWI Hotel} & 720 $\times$ 576	& 389  & 2.5  & Pelligrini et al.\cite{Pellegrini_ICCV_2009} \\
				\rowcolor{green!10} 
				\emph{Crowds Zara} & 720 $\times$ 576 & 204 	& 2.5  & Lerner et al. \cite{Lerner_CGF_2007} \\	
				\rowcolor{green!10} 
				\emph{Crowds Students} & 720 $\times$ 576 & 415  & 2.5  & Lerner et al. \cite{Lerner_CGF_2007} \\	
				\rowcolor{green!10} 
				\emph{Crowds Arxiepiskopi} & 720 $\times$ 576 & 24  & 2.5  & Lerner et al.\cite{Lerner_CGF_2007} \\	
				\rowcolor{green!10} 
				 \emph{PETS 2009} & 768 $\times$ 576 & 19  & 2.5  & Ferryman et al.\cite{Ferryman_PETS_2009}\\	
				\rowcolor{green!10} 
				 \emph{Stanford Drone Dataset (SDD)} & 595 $\times$ 326$*$ & 3295 & 2.5  & Robicquet et al. \cite{Robicquet_ECCV_2016}\\					
					\rowcolor{cyan!10} 
				\emph{BIWI ETH} & 640 $\times$ 480	& 360  & 2.5  & Pelligrini et al.\cite{Pellegrini_ICCV_2009} \\
				\rowcolor{cyan!10} 
				\emph{Crowds Zara} & 720 $\times$ 576 & 148 	& 2.5  & Lerner et al. \cite{Lerner_CGF_2007} \\	
				\rowcolor{cyan!10} 
				\emph{Crowds Uni Examples} & 720 $\times$ 576 & 118  & 2.5  & Lerner et al. \cite{Lerner_CGF_2007} \\	
				\rowcolor{cyan!10} 
				\emph{Stanford Drone Dataset (SDD)} & 595 $\times$ 326$*$ & 3297 & 2.5  & Robicquet et al. \cite{Robicquet_ECCV_2016}\\									
		\hline
\end{tabular}\\
\centering
%Settings: $\sigma^{2}_{v}=2$, $\sigma^{2}_{w}=5$, $t_{update}=3$ \\
%\hspace{}
%\caption{\label{tab:imm_results_2_5_3} Performance summary for the different IMM filter configurations.}
\end{table}

\begin{figure}[h!]
  \begin{center}	
	\begin{tabular}{cc}
				%%\hline
				%%\gradLabel\\
				\includegraphics[width=.4\textwidth,height=0.4\textwidth]{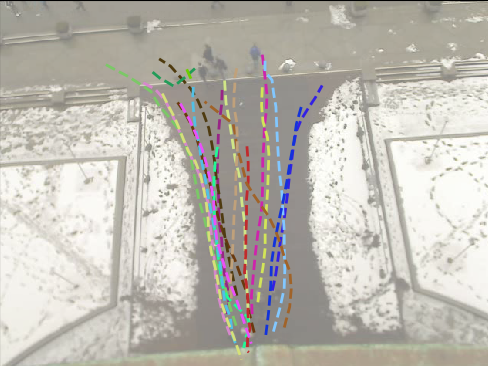} &
				\includegraphics[width=.4\textwidth,height=0.4\textwidth]{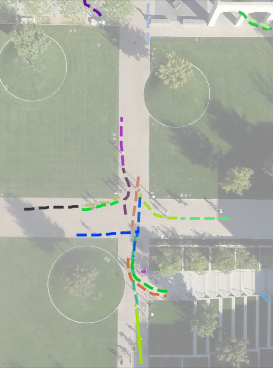}
				\\				
				%%{Example person from INRIA trainig set. }
				%%\hline
	\end{tabular} 
	\end{center}
	\caption{ \label{fig:example_tracks} Example trajectories from the \emph{BIWI ETH} dataset and example tracklets from the sequence \emph{Hyang\_07} from the \emph{Stanford Drone Dataset (SDD)}.} 
\end{figure}

In most datasets, the scene is observed from a bird's eye view, but there are also scenarios where the scene is observed under a higher depression angle. The selected surveillance datasets cover real world scenarios with a varying crowd densities and varying complexity of trajectory patterns. Details of the datasets are summarized in table \ref{tab:dataset_detail} (adapted from \emph{TrajNet} website). The selection includes the following datasets. The \emph{BIWI Walking Pedestrians Dataset} \cite{Pellegrini_ICCV_2009} also sometimes referenced as \emph{ETH Walking Pedestrians (EWAP)}, which is split into two sets (\emph{ETH} and \emph{Hotel}). The \emph{Crowds} dataset also called \emph{UCY "Crowds-by-Example"} dataset \cite{Lerner_CGF_2007} contains three scenes from an oblique view, where the first (Zara) shows a part of a shopping street, the second (\emph{Students}/\emph{Uni Examples}) captures a part of the uni campus and the third scene (\emph{Arxiepiskopi}) captures a different part of the campus. Then, the \emph{Stanford Drone Dataset (SDD)} \cite{Robicquet_ECCV_2016} consists of multiple aerial images capturing different locations around the Stanford campus. And finally the \emph{PETS 2009} dataset \cite{Ferryman_PETS_2009}, where different outdoor crowds activities are observed by multiple static cameras. Sample images with full trajectories and tracklets are shown in figure \ref{fig:example_tracks}.

\begin{figure}[h!]
  \begin{center}	
	\begin{tabular}{cc}
				%%\hline
				%%\gradLabel\\
				\includegraphics[width=.4\textwidth,height=0.4\textwidth]{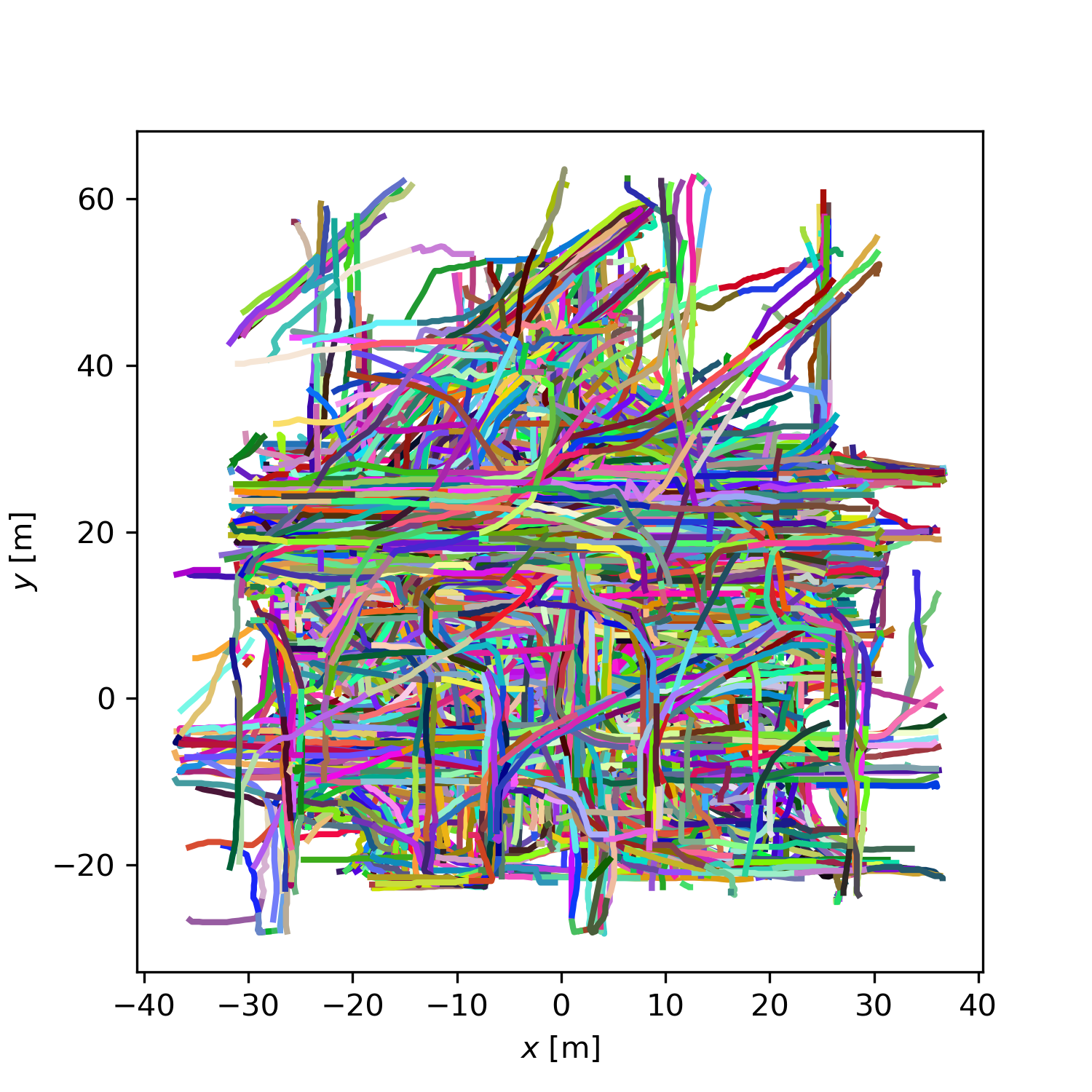} &
				\includegraphics[width=.4\textwidth,height=0.4\textwidth]{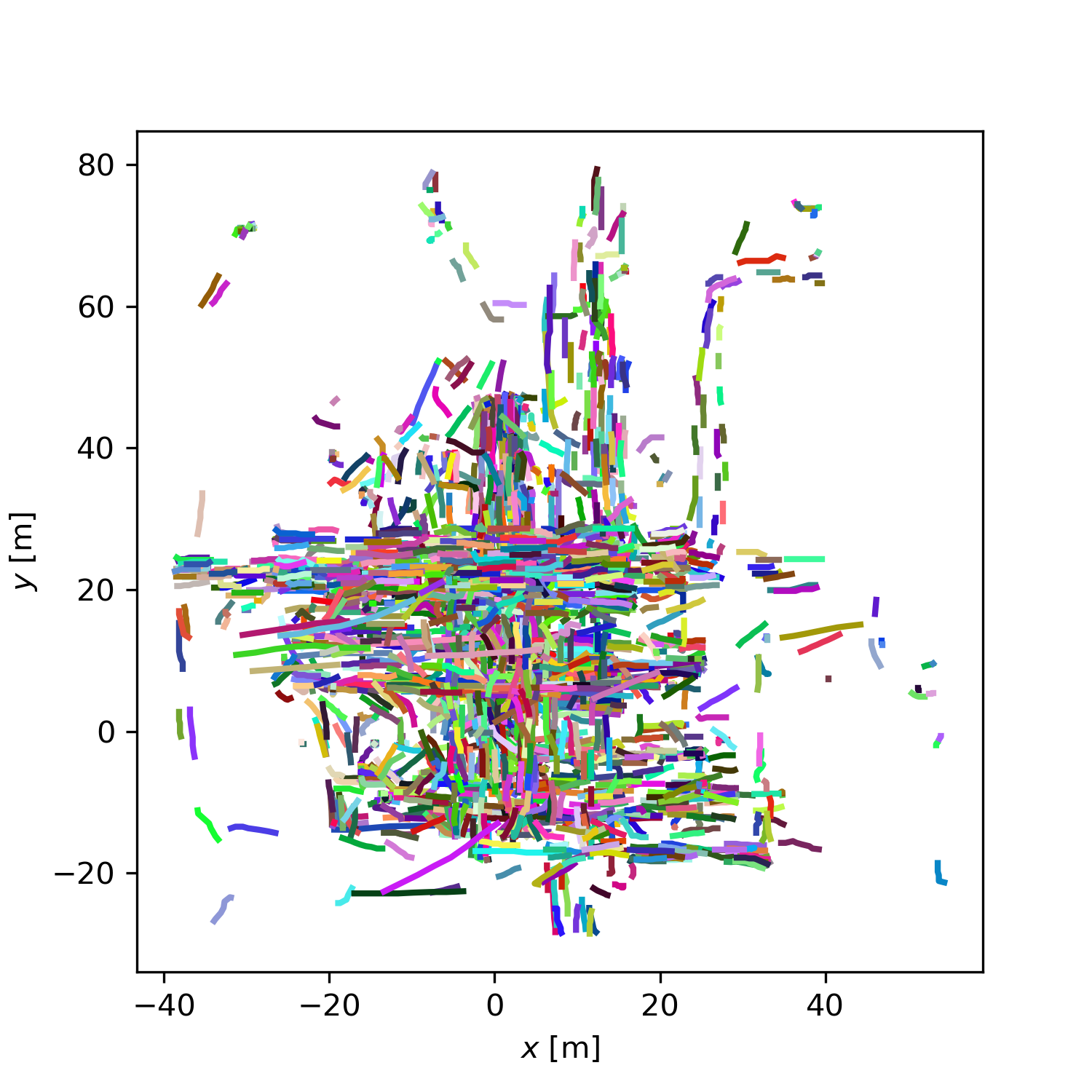}
				\\				
				%%{Example person from INRIA trainig set. }
				%%\hline
	\end{tabular} 
	\end{center}
	\caption{ \label{fig:tracklets_trajnet} (Left) Visualization of all tracklets of the training set from the \emph{TrajNet} dataset collection. (Right) Visualization of all initialization tracklets of the test set.} 
\end{figure}

It is common and good practice to apply cross-validation. For the TajNet challenge, it is done by omitting complete datasets for testing. Because the behavior of humans in crowds is scene-independent and for measuring the generalization capabilities of various approaches across datasets this is very reasonable, in particular for providing a benchmark for human-human interactions. Nevertheless, by combining all training sets the spatial context of scene specific motion and the reference systems are lost. 
When only relying on observed motion trajectories positional information is crucial in order to learn spatio-temporal variation. For example, the sidewalks in the \emph{Hyang} sequences (see figure \ref{fig:example_tracks}) lead to a spatially depending 
change in the curvature of a trajectory. Since our focus is on deep neural networks including RNNs, the shift from position information to higher order motion helps to overcome some drawbacks. Before RNNs were successfully applied for tracking pedestrians in a surveillance scenario, they gained attention due to their success in tasks like speech recognition \cite{Graves_ICASSP_2013,chung2015recurrent} and caption generation \cite{Donahue_CVPR_2015,Xu_MLR_2015}. Since these domain are particularly different to trajectory prediction in certain aspects, their position-dependent movement is not important. Accordingly, RNNs can benefit from conditioning on previous offsets for scene independent motion prediction. 
This insight is not new, yet utilizing offsets really helps not only stabilizing the learning process but also improves the prediction performance for the evaluated networks. This shift to offsets or rather velocities has been also successfully applied for example for the prediction of human poses based on RNNs \cite{Martinez_CVPR_2017}. In the context of deep networks the same effect can also be achieved by adding residual connections, which have been shown to improve performance on deep convolutional networks \cite{He_CVPR_2016}. Presumably due to the limitation of the input and output spaces, for applying on the \emph{TrajNet} challenge instead of prediction of the next position (where will the person be next) predicting the following offsets (where will the person go next) \cite{Hug_RFMI_2017,Hug_arXiv_2018} also contributed to increased prediction accuracy. This becomes immediately apparent by looking at the complete tracklets of the training and test set (see figure \ref{fig:tracklets_trajnet}). 
Firstly, it takes a considerably higher modeling effort to represent all possible positions instead of modeling particular velocities. Further, input data outside the training range can lead to undefined states in the deep network, which result in an unreasonably random output. Some of the initialization tracklets clearly lie outside the training input space. Also, approaches with profit from human-human interaction like \cite{Gupta_CVPR_2018,Hasan_CVPR_2018,alahi2017learning,alahi2016social} in combination with deep networks lack here information about surrounding persons to interact, so that the decoding of relative distances is not possible because of a reduced person density.

\begin{figure}[h!]
  \begin{center}	
	\begin{tabular}{ccc}
				%%\hline
				%%\gradLabel\\
					\includegraphics[width=.3\textwidth,height=.3\textwidth]{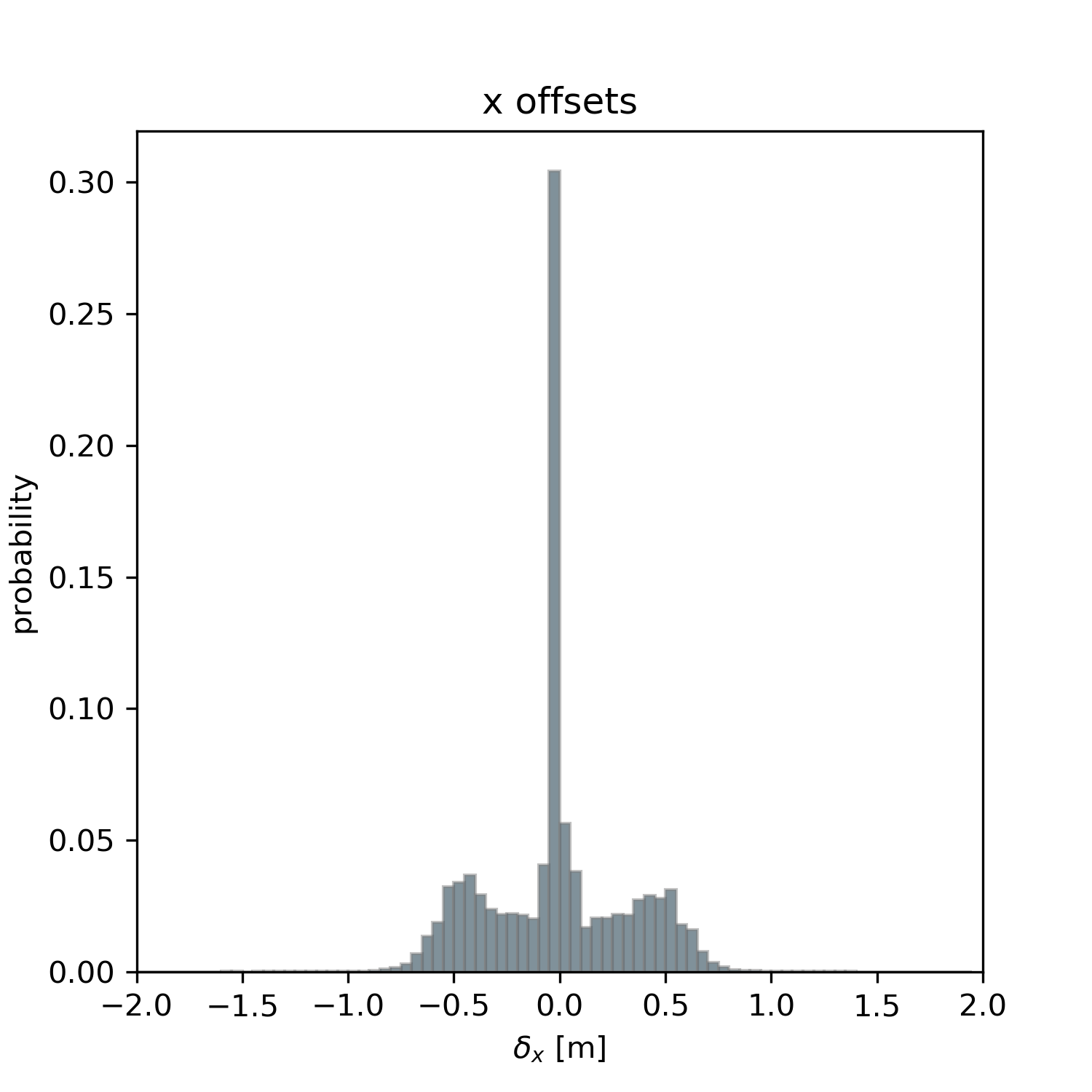} &
				\includegraphics[width=.3\textwidth,height=.3\textwidth]{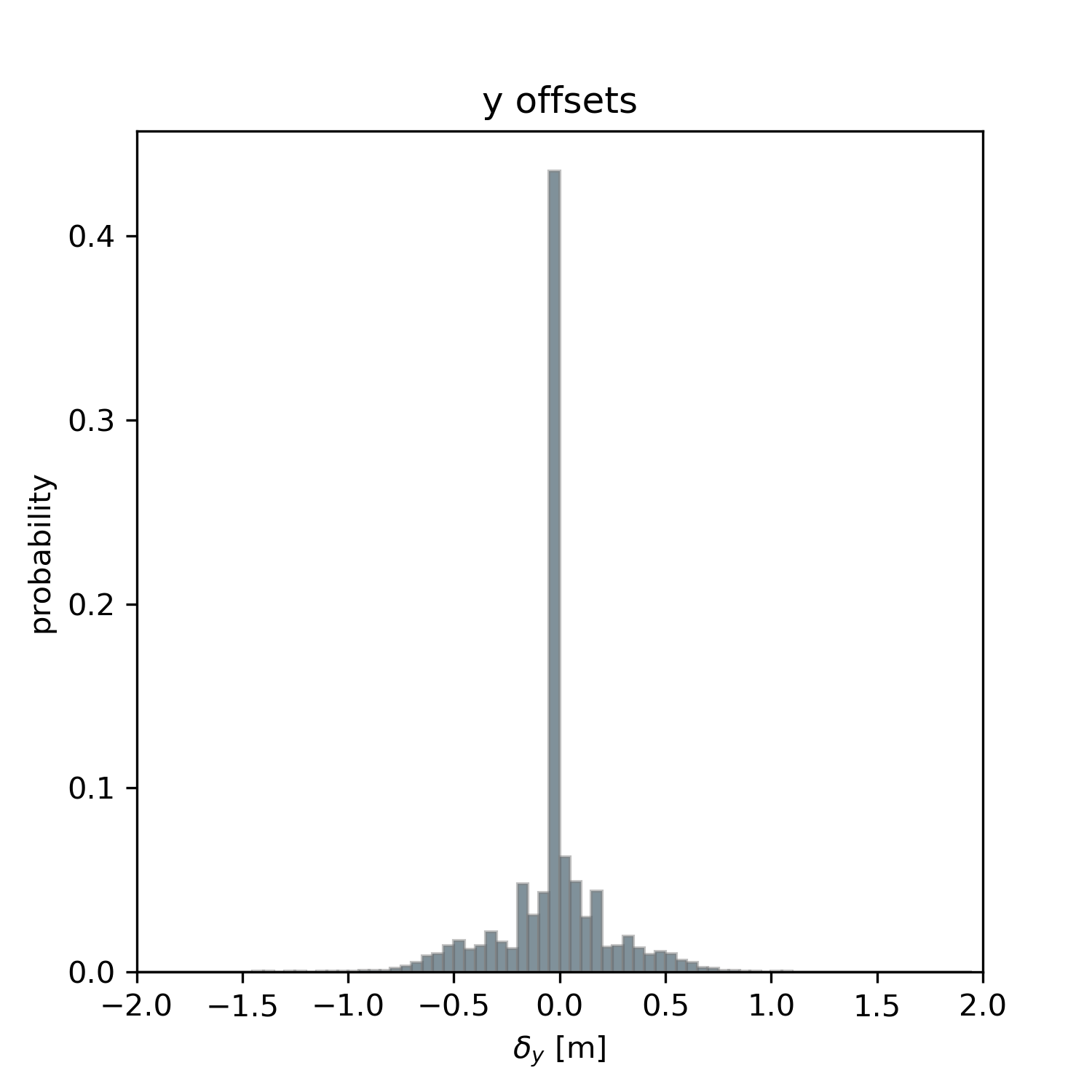} &
				\includegraphics[width=.3\textwidth,height=.3\textwidth]{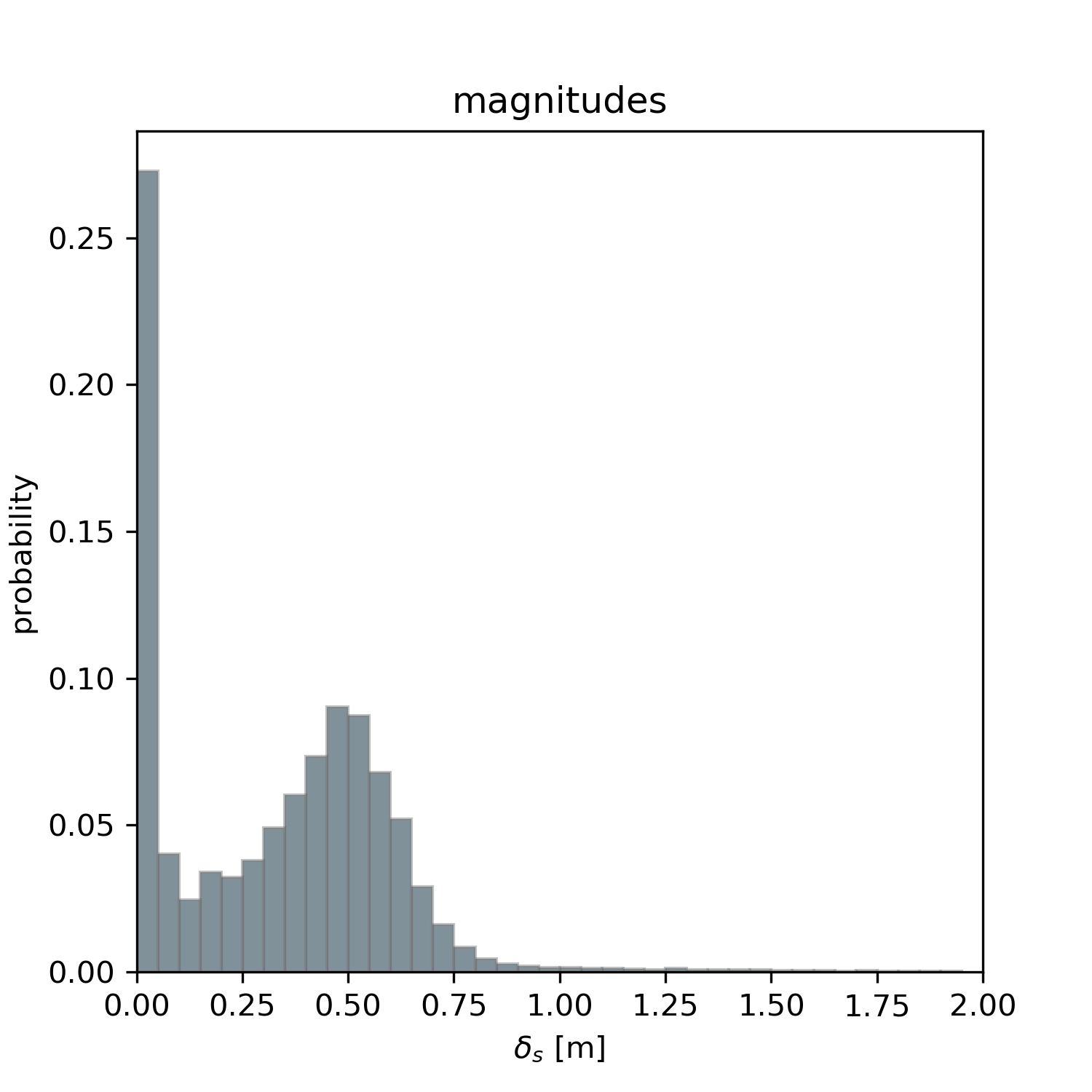}
				\\				
				%%{Example person from INRIA trainig set. }
				%%\hline
	\end{tabular} 
	\end{center}
	\caption{ \label{fig:stats_dataset} (Left, Middle) Offset histograms of the training set. (Right) Magnitude histogram of the offsets.} 
\end{figure}

Another factor for improving the prediction performance is becoming apparent when contemplating the offset distribution of the data. Figure \ref{fig:stats_dataset} shows the offsets histograms for x and y separately. Due to the loss of the reference system, it is impossible to assume a reasonable location distribution a-priori. In contrast, the offset and magnitude distribution clearly reflects the preferred walking speeds in the data. The histograms also show that a large amount of persons is standing. In the recent work of Hasan et al. \cite{Hasan_CVPR_2018}, it was emphasized that forecasting errors are in general higher when the speed of persons is lower and argued that when persons are walking slowly their behavior becomes less predictable, due to physical reasons (less inertia). During our testing we discovered the same phenomenon. In particular RNN based networks tend to overestimate slow velocities and do sometimes not accurately identify the standing behavior. 
Despite this problem, the range of offsets is very limited compared to the location distribution and shows a clear tendency towards expected a-priori values. Common techniques for sequence prediction problems are normalization and standardization of the input data. Whereby normalization has a similar role on the position data, applying standardization on position input data shows no benefit. In our experiments, standardization worked slightly better than normalization or an embedding layer for input encoding. Although the effect on the performance is quite low for the \emph{TrajNet} challenge, our best result is achieved using standardized offsets as input. It is rarely strictly necessary to standardize the inputs, but there are practical reasons like accelerating the training or reducing the chances of getting stuck in local optima \cite{brownlee2017}. Predicting offsets also guarantees that the output directly conforms better with the range of common activation functions.\\

Without discretization artifacts, the dynamic of humans is smooth and persistent. The trajectory data from the \emph{TrajNet} dataset includes varying discretization artifacts or noise levels resulting from different methods with which ground truth data was generated. Part of the ground truth trajectories are generated by a visual tracker or manually annotated. For approximating the amount of noise in the datasets, the distance between a smoothed spline fit through the complete tracklets is compared to the provided ground truth tracklet points. The spline fitting is done with a polynom of degree $k=4$ independent for the x and y values. If the smoothing is too strong, it can drift too far away from the actual data. Nevertheless, the achieved fitted trajectories form a smooth and natural path and are used as rough assessment for the noise levels in the ground truth trajectory data. The results for the training set are summarized in table \ref{tab:dataset_diff_spline}.

\begin{table}[h!]
\caption{Standard deviation of the distance between a smoothed spline fit and the ground truth trajectory data. The average $R^2$ score for all tracklets in the subsets.}
\label{tab:dataset_diff_spline}
 %\centering	
	%\rowcolors{2}{blue!5}{gray!10}
  	\def\arraystretch{1.1}
		\centering
\begin{tabular}{ |c| c c c c |}
		\hline		
		\rowcolor{white!10} 
		\multicolumn{1}{|c|}{Name} & \multicolumn{1}{c}{$\sigma_{x,\text{spline}}$ [m]} & \multicolumn{1}{c}{$\sigma_{y,\text{spline}}$ [m]} &   \multicolumn{1}{c}{$\bar{R^{2}_{x}}$} &  \multicolumn{1}{c|}{$\bar{R^{2}_{y}}$}    \\
				\hline
				\rowcolor{green!10} 
				\emph{Overall} & 0.067	& 0.069 & 0.889 & 0.811  \\
				\hline
				\rowcolor{green!10} 
				\emph{BIWI Hotel} & 0.042	& 0.031 & 0.637 & 0.876  \\
				\rowcolor{green!10} 
				\emph{Crowds Zara\_02} & 0.029 & 0.035 & 0.952 & 0.758 	 \\	
				\rowcolor{green!10} 
				\emph{Crowds Zara\_03} & 0.026 & 0.031 & 0.935 & 0.716 	 \\	
				\rowcolor{green!10} 
				\emph{Crowds Students\_01} & 0.033 & 0.029 & 0.868 & 0.852  \\						
				\rowcolor{green!10} 
				\emph{Crowds Students\_03} & 0.039 & 0.040 & 0.915 & 0.76   \\	
				\rowcolor{green!10} 
				 \emph{Crowds Arxiepiskopi\_01} & 0.050 & 0.027 & 0.959 & 0.677    \\	
				\rowcolor{green!10} 
				 \emph{PETS 2009 S2L1} & 0.037  & 0.026 & 0.781 &  0.877   \\
					\rowcolor{green!10} 
				 \emph{SSD Bookstore\_00} & 0.060 & 0.063 & 0.889 & 0.844   \\
				\rowcolor{green!10} 
				 \emph{SSD Bookstore\_01} & 0.054 & 0.053 & 0.879 & 0.878  \\
				\rowcolor{green!10} 
				 \emph{SSD Bookstore\_02} & 0.068 & 0.073 & 0.861 & 0.921  \\
				\rowcolor{green!10} 
				 \emph{SSD Bookstore\_03} & 0.069 & 0.061 & 0.951 & 0.830  \\
				\rowcolor{green!10} 
				 \emph{SSD Coupa\_03} &  0.057 & 0.043 & 0.954 & 0.937  \\
				\rowcolor{green!10} 
				 \emph{SSD Deathcircle\_00} & 0.072 & 0.079 & 0.893 &  0.808  \\
				\rowcolor{green!10} 
				 \emph{SSD Deathcircle\_01} &0.086  & 0.103 & 0.850 &  0.818   \\
				\rowcolor{green!10} 
				\emph{ SSD Deathcircle\_02} & 0.151 & 0.158 & 0.772 & 0.591 \\
				\rowcolor{green!10} 
				\emph{ SSD Deathcircle\_03} & 0.116 & 0.134 & 0.816 & 0.770 \\
				\rowcolor{green!10} 
				\emph{ SSD Deathcircle\_04} & 0.215  & 0.160 & 0.738 & 0.713 \\
				\rowcolor{green!10} 
				\emph{SSD Gates\_00} & 0.054 & 0.073 & 0.980 & 0.735  \\
				\rowcolor{green!10} 
				\emph{SSD Gates\_01} &0.064 & 0.084 & 0.859 & 0.890  \\
				\rowcolor{green!10} 
				\emph{SSD Gates\_03} & 0.086 & 0.106 & 0.847 & 0.860   \\
				\rowcolor{green!10} 
				\emph{SSD Gates\_04} & 0.071 & 0.155 & 0.820 & 0.906 \\
				\rowcolor{green!10} 
				\emph{SSD Gates\_05} & 0.069 &  0.067 & 0.858 & 0.904  \\
				\rowcolor{green!10} 
				\emph{SSD Gates\_06} & 0.077 & 0.072 & 0.840 & 0.905  \\
				\rowcolor{green!10} 
				\emph{SSD Gates\_07} & 0.084 & 0.126 & 0.908 & 0.817   \\
				\rowcolor{green!10} 
				\emph{SSD Gates\_08} & 0.076 & 0.088 & 0.922 & 0.820   \\				
				\rowcolor{green!10} 
				 \emph{SSD Hyang\_04} & 0.048 & 0.050 & 0.829 & 0.842  \\
				\rowcolor{green!10} 
				 \emph{SSD Hyang\_05} & 0.059 & 0.081 & 0.872 & 0.740  \\
				\rowcolor{green!10} 
				 \emph{SSD Hyang\_06} & 0.070 & 0.066 & 0.875 & 0.811  \\
				\rowcolor{green!10} 
				\emph{SSD Hyang\_07} &  0.040 & 0.079 & 0.879 & 0.894 \\
				\rowcolor{green!10} 
				 \emph{SSD Hyang\_09} & 0.036 &  0.088 & 0.998 & 0.652  \\
				\rowcolor{green!10}
				\emph{SSD Nexus\_00} & 0.076 & 0.082 & 0.886 & 0.742 \\
				\rowcolor{green!10} 
				 \emph{SSD Nexus\_01} &  0.067 &  0.095 & 0.929 & 0.771   \\
				\rowcolor{green!10} 
				 \emph{SSD Nexus\_02} & 0.069 & 0.074 & 0.934 & 0.726   \\
				\rowcolor{green!10} 
				 \emph{SSD Nexus\_03} & 0.188  & 0.113 & 0.786 & 0.572   \\	
				\rowcolor{green!10} 
				 \emph{SSD Nexus\_04} & 0.097  & 0.073 & 0.847  &  0.724   \\	
				 	\rowcolor{green!10} 
				 \emph{SSD Nexus\_07} & 0.053  & 0.069 & 0.935 &  0.764  \\	
					\rowcolor{green!10} 
				 \emph{SSD Nexus\_08} & 0.067  & 0.070  & 0.926 &  0.681  \\
					\rowcolor{green!10} 
				 \emph{SSD Nexus\_09} & 0.052  & 0.094 & 0.913 & 0.816   \\	
		%\hline								
		\hline
\end{tabular}\\
\centering
\end{table}

\begin{figure}[h!]
  \begin{center}	
	\begin{tabular}{cc}
				\includegraphics[width=.4\textwidth]{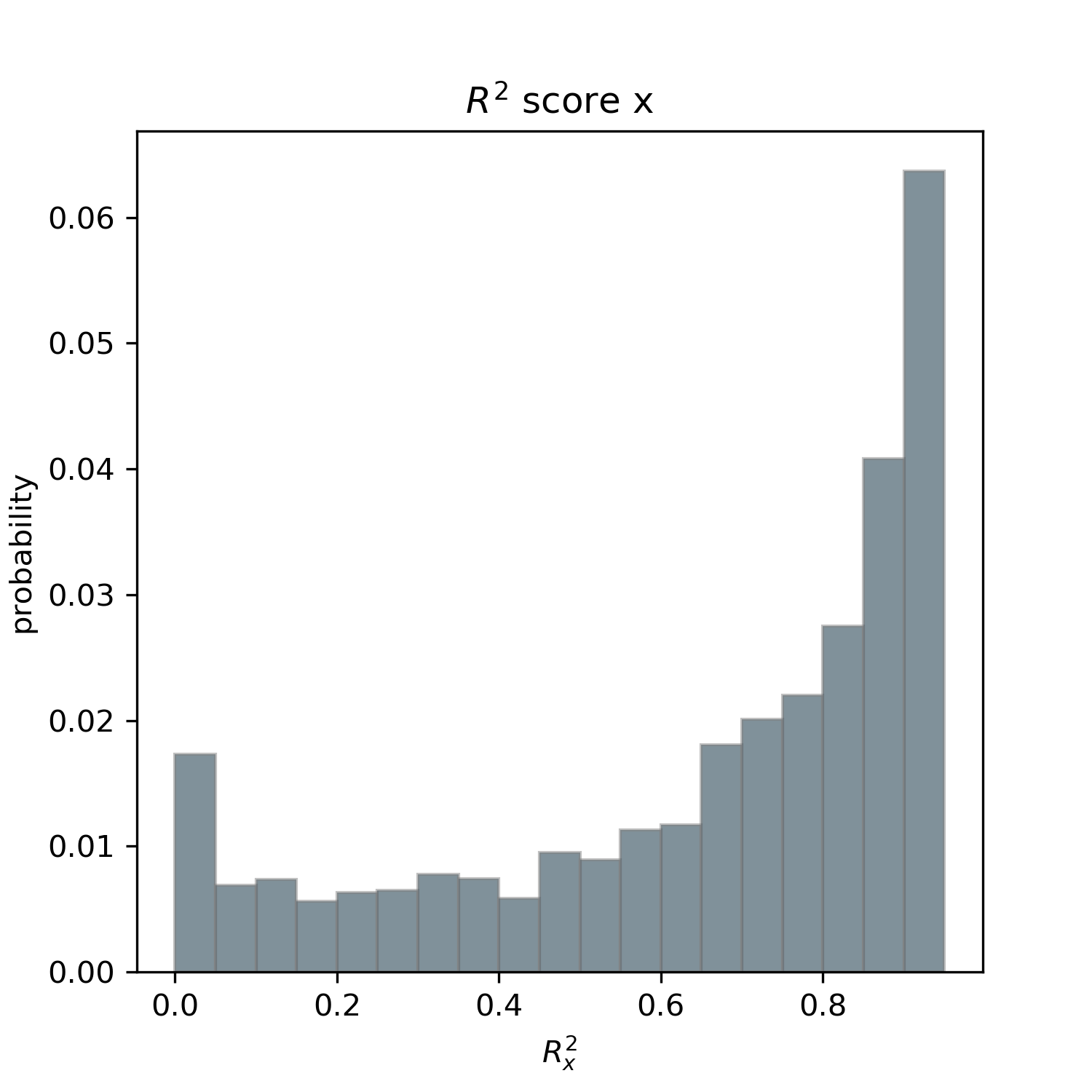} &
				\includegraphics[width=.4\textwidth]{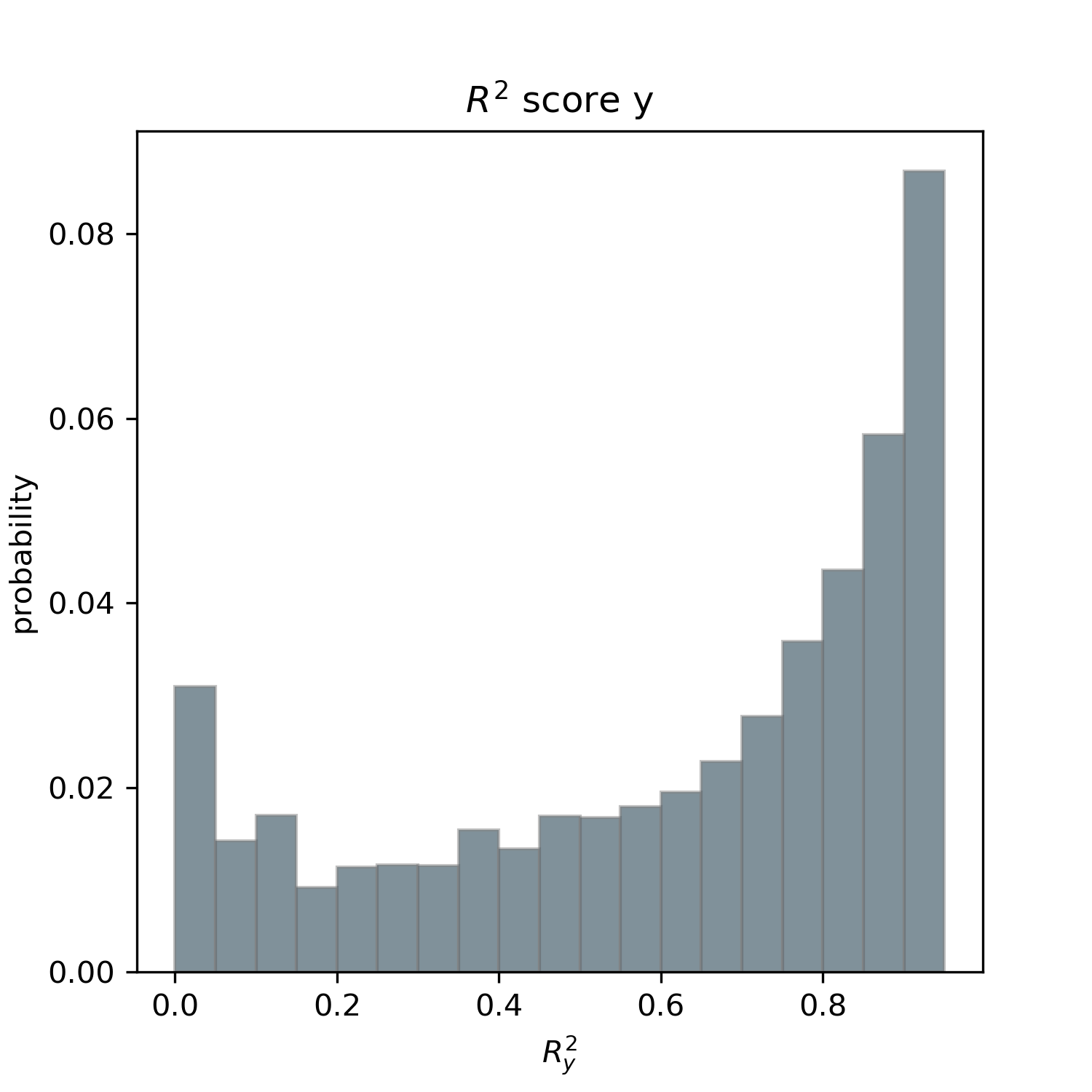}
				\\				
	\end{tabular} 
	\end{center}
	\caption{ \label{fig:r2_coeff} Coefficient of determination $R^2$  for x and y for all training tracklets of the \emph{World H-H TRAJ} challenge. } 
\end{figure}

The approximated noise levels clearly show the variation in the ground truth data. In order to outperform a linear baseline predictor the learned model must be able to successfully model different velocity profiles and capture curved paths out of input data with different noise levels. Due to the varying noise levels, initial experiments to solely train on smoothed fitted trajectories with synthetic noise performed worse. Nevertheless, for the prediction of the future steps the best performing predictor is trained to forecast smoothed paths.\\
Before the different evaluated models are introduced, the last data analysis of the training set is intended to assess the complexity in terms of the non-linearity of the trajectories. Therefore, the coefficient of determination $R^2$ for a linear interpolation is calculated separately for the x and y values. This linear interpolation serves as baseline predictor for the \emph{TrajNet} challenge. The histograms of $R^2$ for the training set are shown in figure \ref{fig:r2_coeff}. $R^2$ is the percentage of the variation that is explained by the model and is used to determine the suitability of the regression fit as a linearity measure \cite{Draper_1966}. The average $R^2$ values are summarized in table \ref{tab:dataset_diff_spline}.
It can be seen that for most tracklets a linear interpolation works very well. In order to outperform the linear interpolation baseline, it is crucial to not only cover a variety of complex observed motions, but to also produce robust results in simpler situations. As mentioned above, the person velocity has to be effectively captured by the model.

\section{Models and Evaluation}
\label{sec:eval}

The goal of this work is by using a sort of coarse to fine searching strategy to reach the maximum achievable prediction accuracy without further cues like human-human interaction or human-space interaction based on basic networks. Towards this end, we started with a set of networks with a limited set of hyper-parameters to narrow it down to one network, in order to then extend the hyper-parameter set for a more exhaustive tuning.
The multi-modal aspect of trajectory prediction is hardly considerable when there is no fixed reference system. Thus, the performance is compared in accordance to the community with the two error metrics of the average displacement error (ADE) and the final displacement error (FDE) (see for example \cite{alahi2016social,Vemula_ICRA_2017,Pellegrini_ICCV_2009,Gupta_CVPR_2018,Xue_WACV_2018,Hasan_CVPR_2018}). The average of both combined values are then used as overall average to rank the approaches. 
The ADE is defined as the average L$2$ distance between ground truth and the prediction over all predicted time steps and the FDE is defined as the L$2$ distance between the predicted final position and the true final position. For the \emph{World H-H TrajNet} challenge the unit of the error metrics is meter. For all experiments, $8$ ($3.2$ seconds) consecutive positions are observed, before predicting the next $12$ ($4.8$ seconds) positions.

Besides the provided approaches of the \emph{World H-H TrajNet} challenge, the following basic neural networks for a coarse evaluation are selected:

\textbf{Multi-Layer-Perceptron (MLP):} The MLP is tested with different linear and non-linear activation functions. One variation concatenates all inputs and predicts $24$ outputs directly. Further, cascaded architectures with a step-wise prediction are examined. We vary between different coordinate system of Euclidean and polar coordinates. As mentioned in section \ref{sec:data_ana}, positions and offsets (also orientation normalized) are considered as inputs and outputs.  

\textbf{RNN-MLP:} RNNs extend feed-forward networks or rather the MLP model due to their recurrent connections between hidden units. Vanilla RNNs produce an output at each time step. For the evaluation of the RNN-MLP, we vary only the MLP layer which is used for the decoding of the positions and offsets. 

\textbf{RNN-Encoder-MLP:} In contrast to the RNN-MLP network, the complete initialization tracklet is used to generate the internal representation before a prediction is done. The RNN-Encoder-MLP is varied by alternating activation functions for the MLP and by alternatively predicting the complete future path/offsets instead of only next steps. As a further alternative, the full path is predicted as offsets to one reference point instead of applying path integration in order to predict the final position.   

\textbf{RNN-Encoder-Decoder-Model (Seq$2$Seq):} In addition to RNN-Encoder-MLPs, Seq$2$Seqs include a second network. This second decoder network takes the internal representation of the encoder and then starts predicting the next steps. The different settings for the evaluation of this model where due to alternating activation functions for the MLP on top of the decoder RNN.   

\textbf{Temporal Convolutional Networks (TCN):} As an alternative to RNNs and based on \emph{WaveNets}\cite{Oord_arXiv_2016}, Bai et al. \cite{Bai_arXiv_2018} introduced a general convolution architecture for sequence prediction. We tested their standard and extended architecture with a gating mechanism (GTCN). For a more detailed description, we refer to the original papers. 

All networks were trained with varying number of layers ($1$ to $5$) and hidden units ($4$ to $64$) using stochastic gradient descent with a fixed learning rate of $0.005$. The models are trained for $100$ epochs using ADAM optimizer \cite{Kingma_ICLR_2015} and have been implemented in \emph{Tensorflow} \cite{tensorflow}. Firstly, only standard RNN cells are used for the experiments. Later, we also tested with RNNs variants Long Short-Term Memory \cite{Hochreiter_NC_1997} (LSTM) and Gated Recurrent Unit \cite{Cho_EMLN_2014} (GRU). As loss the mean squared error between the predicted and the ground truth position or offsets over all time steps is used.\\ 
In order to emphasize trends a part from the result of the first experiments are summarized in table \ref{tab:Trajnet_challenge_results} (highlighted in gray). The best results were achieved with the RNN-Encoder-MLP. However, in most cases the different architectures perform very similar. These initial result also show that the best performing networks lie close to the result achieved with linear interpolation. Outlier weak performances are due some strong overestimation of slow person velocities and some undefined random predictions when using positions. Hasan et al. reduced this effect by integrating head pose information. We can only remark for the tested networks that this effect can also differ for different runs. Naturally it is important that during training the networks see enough samples from standing of slow moving situations. Excluding such samples through heuristic or probabilistic filtering only helps during application.\\

\begin{table}[h!]
\caption{Results for the world plane human-human dataset challenge (\emph{World H-H TRAJ} challenge).}
\label{tab:Trajnet_challenge_results}
 %\centering	
\rowcolors{2}{blue!5}{gray!10}
\def\arraystretch{1.1}
\hspace*{-0cm}\begin{tabular}{ |c| c c c c |}
		\hline					
			\rowcolor{white!10} 
		\multicolumn{1}{|c|}{Approach} & \multicolumn{1}{c}{Overall Average $\downarrow$} & \multicolumn{1}{c}{FDE [m] $\downarrow$} & \multicolumn{1}{c}{ADE [m] $\downarrow$} & \multicolumn{1}{c|}{Reference}\\
				\hline
				\rowcolor{red!10}
				RED & \textbf{0.797} &	1.229 & 0.364 & Ours \\
				\rowcolor{blue!10}				
				Social Forces (EWAP) & \textbf{0.819}  & 1.266 & 0.371  & Helbing and Moln\'ar \cite{Helbing_PRE_1995} \\
				\rowcolor{blue!10}
				Social Forces (ATTR) & 0.904 & 1.395 & 0.412 & Helbing and Moln\'ar \cite{Helbing_PRE_1995} \\
				\rowcolor{blue!10}
				social lstm\_v2	&	1.387 &	2.098	& 0.675	&	Alahi et al.\cite{alahi2016social} \\
				\rowcolor{blue!10}
				social lstm	&	1.563 &	2.299	& 0.826	& Alahi et al.\cite{alahi2016social} \\
				\rowcolor{blue!10}
				social lstm\_v3 & 2.874	&	4.323 &		1.424 & Alahi et al.\cite{alahi2016social} \\
				\rowcolor{blue!10}
				Interactive Gaussian Processes &	1.642 &	1.038 &	2.245	&	Ellis et al. \cite{Ellis_ICCVW_2009} \\
				\rowcolor{yellow!10}
				Linear Interpolation	&	0.894	&1.359	& 0.429	&	 \\
				\rowcolor{gray!10}
				Linear MLP (Pos) & 1.041 & 1.592 & 0.491 & \\
				\rowcolor{gray!10}
				Linear MLP (Off) & 0.896  & 1.384 & 0.407 &\\
				\rowcolor{gray!10}
				Non-Linear MLP (Off) & 2.103 & 3.181 & 	1.024 &  \\	
				\rowcolor{gray!10}
				Linear RNN & 0.951 & 1.482 & 0.420 & \\
				\rowcolor{gray!10}
				Non-Linear RNN & \textbf{0.841} & 1.300 & 0.381 & \\
				\rowcolor{gray!10}
				Linear RNN-Encoder-MLP & 0.892 & 1.381 & 0.404 & \\
				\rowcolor{gray!10}
				Non-Linear RNN-Encoder-MLP & \textbf{0.827} & 1.276 & 0.377 & \\
				\rowcolor{gray!10}
        Linear Seq2Seq & 0.923 & 1.429 & 0.418 & \\
					\rowcolor{gray!10}
        Non-Linear Seq2Seq & \textbf{0.860} & 1.331 & 0.390 & \\
					\rowcolor{gray!10}
        TCN & \textbf{0.841} & 1.301  & 0.381 &  Bai et al. \cite{Bai_arXiv_2018}\\
					\rowcolor{gray!10}
        Gated TCN & 0.947 & 1.468 & 0.426 & Bai et al.\cite{Bai_arXiv_2018} \\					
		\hline
\end{tabular}\\
\centering
Results highlighted in blue are taken from the \emph{TrajNet} website \cite{TrajNet_2018} \\ (\url{http://trajnet.stanford.edu/}, accessed 19.05.2018)\\
\end{table}

There is no network that is clearly performing best, thus the gap between a MLP predictor and a Seq$2$Seq model is very narrow in the test scenarios. However, besides the factors derived from the data analysis, a prediction of the full path instead of step-wise prediction helps to overcome an accumulation of errors that are fed back into the networks. For the \emph{TrajNet} challenge with a fixed prediction horizon, we thus prefer the RNN-Encoder-MLP over a Seq$2$Seq model. In the domain of human pose prediction based on RNNs, Li et al \cite{Li_arXiv_2017} reduced this problem with an Auto-Conditioned RNN Network and Martinez et al. \cite{Martinez_CVPR_2017} propose using a Seq$2$Seq model along with a sampling-based loss. The TCNs perform here similar to RNNs. Since RNNs are more common, also as part of architectures which model interactions (see \cite{alahi2016social,alahi2017learning,Hasan_CVPR_2018,Xue_WACV_2018}) to represent single motion, we keep the RNN-Encoder-MLP as our favored model.\\
\textbf{RNN-Encoder-MLP $\rightarrow$ RED-predictor:}
According to the training set analysis and the comparison of architectures the selected model for the \emph{TrajNet} challenge modeling only single human motion is a RNN-Encoder-MLP. The RNN-Encoder can generalize to deal with varying noisy inputs and thus is able to better capture the person motion compared to the linear interpolation baseline. The main insight is that motion continuity is easier to express in offsets or velocities, because it takes considerably more modeling effort to represent all possible conditioning positions. Especially for the \emph{World H-H TRAJ} challenge, with the different range for positions in the training and test set, this has significant influence on whether a good performance can be obtained.
In combination with easier, but helpful, data-prepossessing, the transfer to directly using smoothed trajectories as desired output, and a full path prediction to prevent error accumulation during a step-wise prediction, our simple but effective baseline predictor for the \emph{TrajNet} challenge is ready. Recurrent-Encoder with a dense MLP layer stacked on top the predictor is referred to as RED-predictor. As reference point for all predicted offsets of the full smooth path the last input position is used. Full path integration worked similar well, but here offsets to the reference positions are predicted\\
The best achieved result is highlighted in red in table \ref{tab:Trajnet_challenge_results}. After a fine search for this network, the shown result is produced with a LSTM cell (state size of $32$) and one recurrent layer. The proposed predictor was able to produce sophisticated results compared to elaborated models which additionally rely on interaction information like the model from Helbing and Moln\'ar \cite{Helbing_PRE_1995} and the Social-LSTM \cite{alahi2016social}.  
The Social-LSTM is one of the first proposed RNN-based architectures which includes human-human interaction and laid the basis for architectures like presented in the work of Hasan et al.\cite{Hasan_CVPR_2018} or Xue et al. \cite{Xue_WACV_2018}. Single motion is modeled with an LSTM network. By applying some of the proposed factors to the model, it is expected that the model and equity accordingly model extensions are able to outperform the proposed single motion predictor.

\section{Discussion and Failure Cases}
\label{sec:discussion}
\begin{figure}[h!]
  \begin{center}	
	\begin{tabular}{cc}
				\includegraphics[height=.4\textwidth]{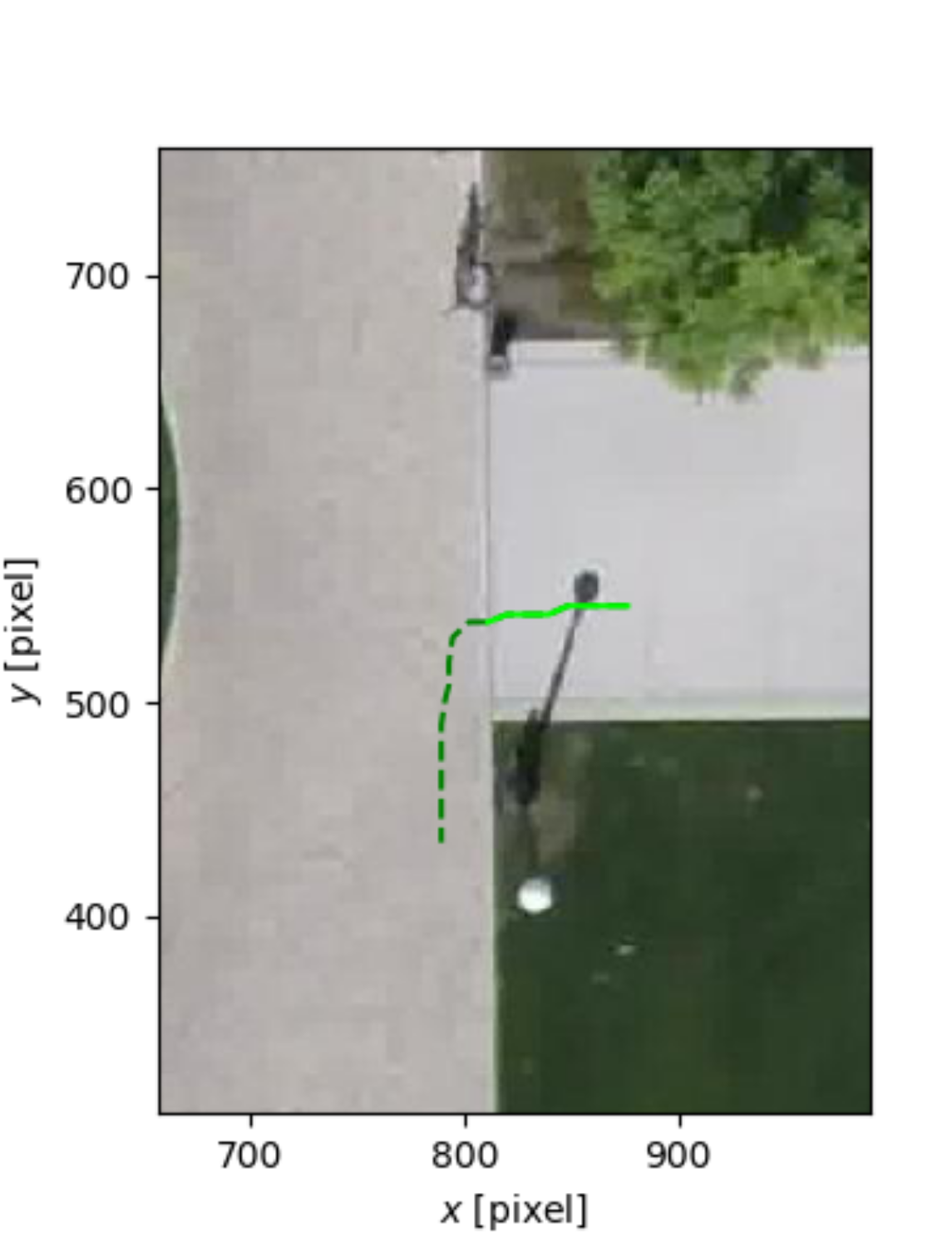} &
				\includegraphics[height=.4\textwidth]{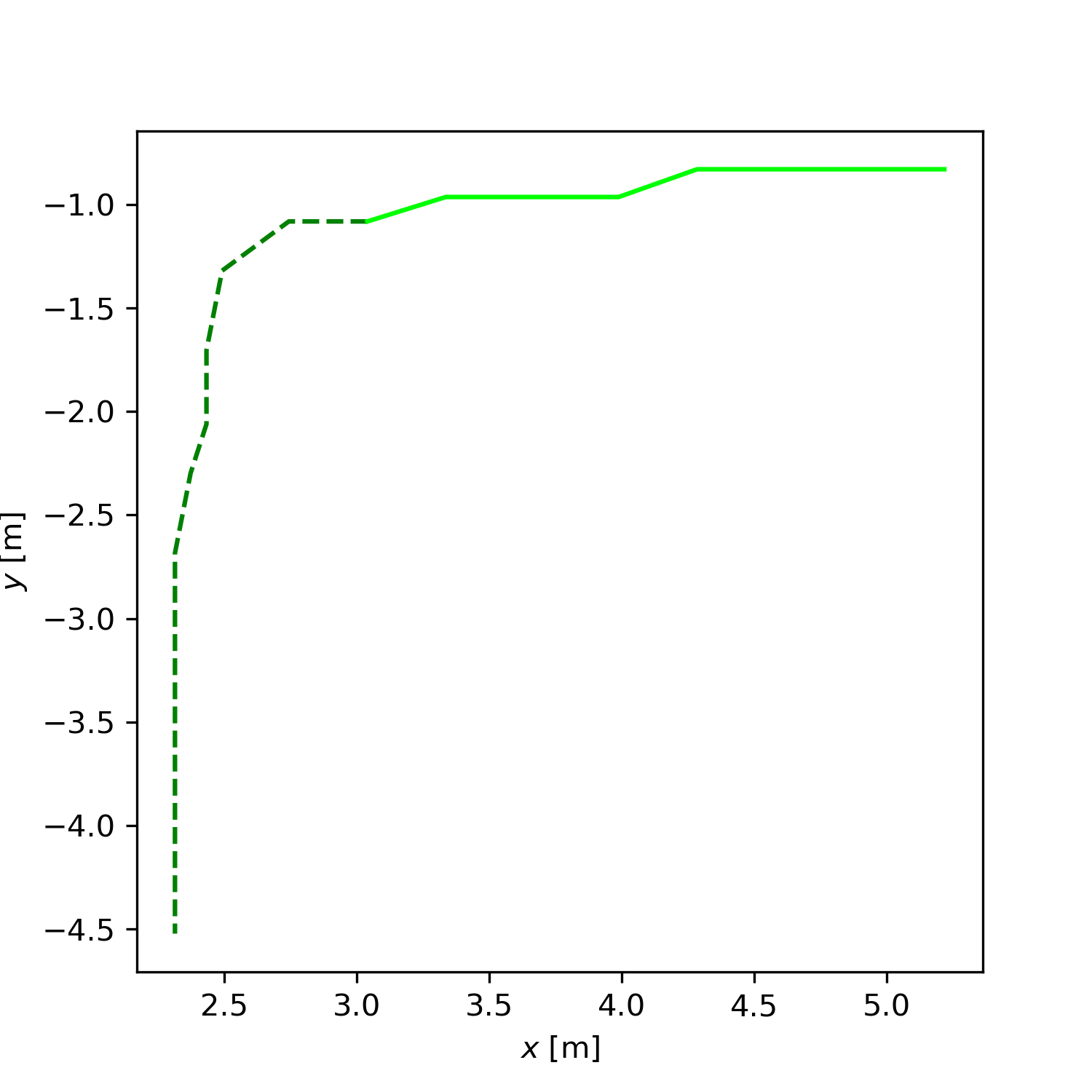}
				\\				
	\end{tabular} 
	\end{center}
	\caption{ \label{fig:intersection} Example where the scene context strongly influences the person trajectory. The initialization tracklet (solid line) delivers no evidence for a turning maneuver at the intersection. This also shows the multi-modal nature of the prediction problem. } 
\end{figure}

After emphasizing the factors needed in order to achieve sophisticated results based on standard neural networks in the above sections, in this section we discuss some failure cases.\\
Without exploiting scene-specific knowledge for trajectory prediction, some particular changing behavior in the human motion is not predictable. For example, in the shown tracklet from \emph{SSD Hyang} (see figure \ref{fig:intersection}), there is no cue for a turning maneuver in the initialization tracklet. In order to correct the prediction, new observations are required. 
All methods tend to predict in such a situation a relatively straight line, resulting in a high prediction error. A scene-independent motion representation is pursuant to better generalize, but for overcoming some limitation in the achievable prediction accuracy, the spatial context is required. The sample tracklet also illustrates the multi-modal nature of the prediction problem. While the person is making a left turn, it is also possible to make a right turn. By using a single maximum-likelihood path the multi-modality of a motion and the uncertainty in the prediction is not covered. The prediction uncertainty can be considered by using the normalized estimation error square (nees) \cite{Huber_Phd_2015}, also known as Mahalanobis distance, which corresponds to a weighted Euclidean distance of the errors. But most methods are designed as a regression model, thus for a unified evaluation system the Mahalanobis distance is not applicable. As mentioned, there are a few approaches which include the multi-modal aspect of the problem \cite{Kitani_ECCV_2012,Lee_CVPR_2017,Hug_arXiv_2018}. Without additional cues of the current scene, these approaches are limited to a fixed scene.\\

Independent of the question how to include all aspects of a problem in a unified benchmarking, they strongly influence the possible achievable results. The results presented in section \ref{sec:eval} show that independent from the model complexity approaches restricted to observing only information from one trajectory are in range to their reachable performance limit on the current dataset repository. Of course due to the fast development in the field of deep neural networks there is still space for improvement, but the current benchmark cannot be completely solved. However, the \emph{TrajNet} challenges also provides human-human and human-space information and recent work like the approaches of Gupta et al. \cite{Gupta_CVPR_2018}(human-human) or Xua et al. \cite{Xue_WACV_2018} and Sadeghian et al. \cite{Sadeghian_arXiv_2018a} (human-human, human-space) show possibilities of how to further improve the performance accuracy.\\

\section{Conclusion}
\label{sec:conclusion}
In this paper, we presented an evaluation of deep learning approaches for trajectory prediction on \emph{TrajNet} benchmark dataset. The initial results showed that without further cues like human-human interaction or human-space interaction most basic networks achieve similar results in small range close to a maximum achievable prediction accuracy. By modifying a standard RNN prediction model, we were able to provide a simple but effective RNN architecture that achieves a performance comparable to more elaborated models. 

\textbf{Acknowledgements :} The authors thank the organizers of the \emph{TrajNet} challenge for providing a framework towards a more meaningful, standardized trajectory prediction benchmarking.

%%%%%%%%%%%%%%%%%%%%%%%%%%%%%%%%%%%%%%%%%%%%%%%%%%%%%%%%%%%%%
%%%%% References %%%%%
%TODO gray/grey and color/colour (also in plots!)
%TODO references: make shure to write dollar with á
%TODO delete doto references
\bibliography{TrajNet}   %>>>> bibliography data in report.bib
\bibliographystyle{spiebib}   %>>>> makes bibtex use spiebib.bst
%%%%%%%%%%%%%%%%%%%%%%%%%%%%%%%%%%%%%%%%%%%%%%%%%%%%%%%%%%%%%%

\end{document}